# Alternative Techniques for Mapping Paths to HLAI

**Ross Gruetzemacher, David Paradice**
Harbert College of Business, Department of Systems and Technology, Auburn University
rossg@auburn.edu, dparadice@auburn.edu

## Abstract

The only systematic mapping of the HLAI technical landscape was conducted at a workshop in 2009 [Adams et al., 2012]. However, the results from it were not what organizers had hoped for [Goertzel 2014, 2016], merely just a series of milestones, up to 50% of which could be argued to have been completed already. We consider two more recent articles outlining paths to human-like intelligence [Mikolov et al., 2016; Lake et al., 2017]. These offer technical and more refined assessments of the requirements for HLAI rather than just milestones. While useful, they also have limitations. To address these limitations we propose the use of alternative techniques for an updated systematic mapping of the paths to HLAI. The newly proposed alternative techniques can model complex paths of future technologies using intricate directed graphs. Specifically, there are two classes of alternative techniques that we consider: scenario mapping methods and techniques for eliciting expert opinion through digital platforms and crowdsourcing. We assess the viability and utility of both the previous and alternative techniques, finding that the proposed alternative techniques could be very beneficial in advancing the existing body of knowledge on the plausible frameworks for creating HLAI. In conclusion, we encourage discussion and debate to initiate efforts to use these proposed techniques for mapping paths to HLAI.

## 1 Introduction

Human-level artificial intelligence (HLAI) is thought by many to be the grand objective of AI, akin to the field's theory of everything. To be specific, a substantial portion of the field would disagree with this – there is still not even consensus on whether it is possible – but, despite the lack of consensus, interest in it has increased significantly in the past decade. To our knowledge only two maps or roadmaps for HLAI have been proposed; one systematic mapping of the HLAI technical landscape [Adams et al., 2012] and one roadmap for machine intelligence [Mikolov et al., 2016]. However, Lake et al. [2017] have carefully analyzed the development of machines that think and learn like humans, and although not a mapping per se, we feel their work is worthy of consideration in this context as well.

The paper will proceed by first examining the three relevant studies mentioned. Initially, we focus on the mapping of the HLAI technical landscape from the 2009 workshop. This was the first and perhaps only collective attempt since Minsky et al. [2004] to try and identify a common path forward for creating HLAI. The result was successful in that it offered rough milestones that could qualitatively be assessed, but fell short of organizers expectations [Goertzel 2014, 2016]. We then explore the work of Mikolov et al. [2016] and Lake et al. [2017] on possible paths for developing HLAI. Each of these studies offers a different perspective that may help to better inform future efforts to create an updated HLAI mapping. We then consider methods; first, we consider technological roadmapping, which was used in the 2009 workshop, then we consider alternative mapping methods from scenario planning literature as well as the possibilities for using digital platforms and crowdsourcing. A discussion ensues where we analyze the previous studies relative to recent progress, examine the existing work and alternative mapping techniques, and consider the need for an updated systematic mapping of the HLAI. Finally, we conclude by encouraging discussion and debate over initiating efforts to map the paths to HLAI.

## 2 Background

Before we discuss the previous studies, we note that a true mapping of the HLAI technical landscape, in the sense of a directed graph, has never been conducted. We believe this to be a significant challenge, but also a worthy objective in the value it could provide the field. In this section we describe the existing outlines and mappings for developing HLAI.

**2.1 Mapping of the HLAI Technical Landscape**

The only systematic mapping of the HLAI technical landscape was conducted at a small workshop in 2009 [Adams et al., 2012], and much progress has been made in the field of AI in the decade since. Particularly, deep learning, deep reinforcement learning and natural language processing have seen significant milestones achieved [Mnih et al., 2015; Radford et al. 2019; Silver et al., 2017]. Considering these advances over the past decade, we believe that it is time to consider an updated mapping of HLAI.

The 2009 workshop was intended to create a roadmap to HLAI by "crafting a series of milestones, measuring step-by-step progress along an agreed upon path from the present state of AI to human-level AGI" [Goertzel 2014]. However, this was more difficult than had been anticipated. Only 13 experts participated, all of whom were preselected because their primary research focus was HLAI. Ultimately, organizers found it difficult to get a consensus on much at all – researchers tended to advocate roadmaps that were consistent with their own research – and the roadmap did not materialize [Goertzel 2016]. Due to these challenges, the roadmap turned out more like a mapping, and the organizers concluded that getting to HLAI was similar to climbing a mountain, with numerous possible paths, the easiest of which is difficult to tell.

The ideas that were explored in the workshop built on the results from two previous workshops conducted in 2008 and 2009 on "Evaluation and Metrics for Human-Level AI" [Laird 2009; Goertzel 2014]. Specifically, the starting point for the workshop was a slightly modified version of the architecture proposed by Laird and Wray [2010]. This included characteristics for HLAI environments, tasks and agents, as well as requirements for cognitive architectures for HLAI.

The results of the workshop were a loosely bound map of scenario milestones moving generally toward HLAI that were laid out on a grid with axes corresponding to the cognitive development theories of Piaget [1964] and Vygotsky [1978]. Seven scenarios were posited, each of which was representative of a combination of tasks that were to be performed within a specified environment [Adams et al., 2012]. The scenarios were laid out in such a way that progress toward the completion of these scenarios was representative of different types of progress in the conjoined cognitive development model. A reproduction of the mapping from the 2009 workshop is shown in Figure 1 in the top right corner.

The seven scenarios that were posited were general video-game learning (GVL), scene or story comprehension, preschool learning, reading comprehension, school learning, the Wozniak test [Moon 2007] and the Nilsson general employment test for HLAI [Nilsson 2005; Adams et al., 2012]. As can be seen in Figure 1, GVL is associated with all levels of cognitive development. Preschool learning is associated with lower middle levels of cognitive development. Scene and story comprehension is associated with lower middle to high levels of cognitive development. Reading comprehension is associated with middle to high levels of cognitive development and school learning is associated with similar levels of cognitive development but with higher requirements for tool integration. The Wozniak test is associated with high levels of cognitive development, but still falls short of complete human-level intelligence, while the Nilsson test requires the highest levels of cognitive development.

### 2.2 Roadmap to Machine Intelligence

Mikolov et al. [2016] have reported a roadmap for machine intelligence that offers a different approach to the notion of mapping the path to a general form of artificial intelligence. While it offers a the description of a full training environment and routine for the development of HLAI, it is also incomplete in that it does not offer concrete proposals for the more challenging tasks required in the development process. However, it does represent the only proposed end-to-end description of a process for training an HLAI agent.

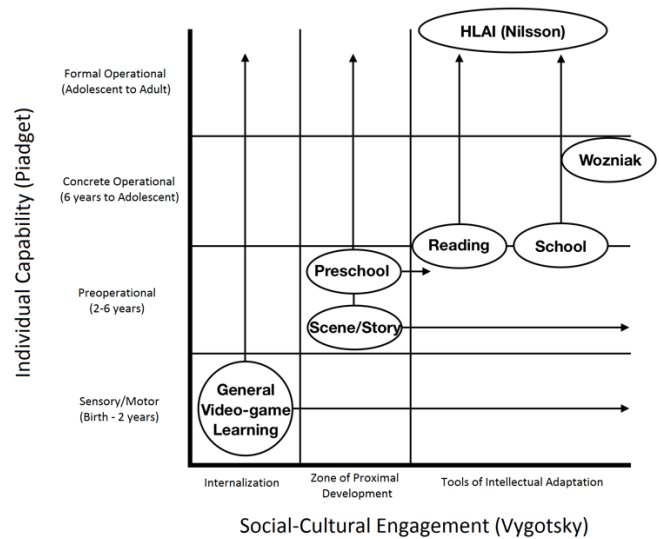

**Figure 1**: A depiction of the mapping of the HLAI technical landscape from 2009 [Adams et al., 2012].

Mikolov et al. [2016] believe that progress can best be made by realistic, small, incremental steps that can jointly lead us to the ultimate goal of creating and implementing a machine that can *communicate* and *learn* new concepts at a pace similar to that of a human. Communication and learning are central to their formulation of an intelligent machine, and they go on to propose them as formal desiderata for their roadmap. Communication is critical because it is an interactive means for transferring information, and it is thus necessary for transferring knowledge from humans, in the form of commands or instructions, and for receiving feedback from machines. They further propose natural language to be the general interface for intelligent machines. Learning is necessary for a machine to adapt and modify itself based on experience, for example, it must learn to correct its mistakes. Here, they also note that machines will start as life as a *tabula rasa*, and will have to learn everything from nothing. While this learning may be child-like in some ways, they are careful not to anthropomorphize the learning process.

They go on to suggest the use of a simulated ecosystem for educating communication-based intelligent machines [Mikolov et al., 2016]. The proposed ecosystem is suggested to be like a "kindergarten" that teaches the machine the basics of linguistic interaction and how to operate in the real world. They suggest that the environment should be challenging enough that it requires meta learning, yet manageable enough that a human could navigate even if communicating in an unknown language. The ecosystem would contain two agents: a learner and a teacher. The teacher would assign tasks and reward the learner for desirable behavior, but the teacher's behavior would be entirely scripted by the developers. They then outline a structured process by which the learner can be educated in the environment, and finally, they propose a scenario in which the trained agent may interact in a real-world situation following training.

What has been described in the previous two paragraphs can be thought of as what Mikolov et al. [2016] propose as a roadmap to machine intelligence. It is most similar in nature to the notion of a continual learning agent set forth by Ring [1997], who describes an agent with experiences that "occur sequentially, and what it learns at one time step while solving one task, it can use later, perhaps to solve a completely different task.". However, it is devoid of concrete suggestions for some other critical elements to an intelligent machine that they describe. Such elements include the mastery of different types of learning, the need for long-term memory capacity and the use of compositional learning skills to enable learning without any new information. They conclude by comparing their proposed roadmap with other related work.

### 2.3 Creating Human-like Machines

Lake et al. [2017] present yet another approach to describing the paths toward the development of human-like machines, which, although not a roadmap per se, is a comprehensive documentation of likely elements of HLAI. Their work differs from Mikolov et al. [2016] in a significant sense because it focuses specifically on brain-inspired human-like intelligence, but in other ways both proposals are very similar. A primary similarity is that they both place significant emphasis on learning, especially meta learning and compositionality. However, a notable difference is that Lake et al. discuss the necessity of reinforcement learning, particularly the model-based variant, whereas the proposed method of Mikolov et al., while resembling reinforcement learning in many ways, differs markedly in others due to their emphasis on natural language communication.

Lake et al. [2017] use the paper to propose three core sets of ingredients for building more human-like learning and thinking machines. We believe that in this way, it resembles the notion of a mapping or roadmap. We describe the three sets of core ingredients in the following paragraphs.

The first of these sets is focused on the developmental "start-up software" of agents, i.e. cognitive capabilities that are present early in development [Lake et al., 2017]. In this group they focus on two primary components: intuitive physics and intuitive psychology. Intuitive physics refers to infants' primitive object concepts that allow for understanding the physical world. Intuitive psychology refers to the fact that infants have an innate or rapidly learned theory of mind, which enables them to attribute thoughts and agency to themselves and other people or animals. These cognitive capabilities are natural for humans and enable us to build predictive models and make inferences that accelerate learning.

The second set of these ingredients concerns learning, and focuses on three primary ingredients: compositionality, causality and meta learning [Lake et al., 2017]. Compositionality is the idea that new representations are able to be constructed by combining primitive elements, e.g. understanding how pen strokes combine to form characters or how sub-goals can combine forming goals. For scene understanding and concept learning, causal models are representative of real-world processes that create perceptual observations. In reinforcement learning and control, causal models are used to represent the structure of the environment, e.g. models for state-to-state transitions or action/state-to-state transitions. Meta learning, or "learning-to-learn," is closely associated with transfer learning, multitask learning and representation learning, terms that refer to ways in which learning new tasks or concepts can be accelerated through previous or parallel learning. Lake et al. see these ingredients as essential to machines' ability to learn rich models from sparse data like humans.

The final of the sets of ingredients concerns the speed with which these rich models can be used for inference [Lake et al., 2017]. Here, they discuss popular algorithms for approximate inference in structured models, including Monte Carlo based techniques. However, they also discuss shortcomings of such techniques in certain applications and domains. They then take up a discussion of model-based and model-free reinforcement learning, noting that the brain likely uses both. Model-based learning in the brain is used for planning action sequences for more complex tasks by building a "cognitive map." Oddly enough, a scenario mapping technique we will discuss in the following section is inspired by cognitive maps.

Prior to the conclusion, there is also a section about looking forward which discusses future possibilities for deep learning, including things like attention and memory. The significance of some things mentioned in this discussion was confirmed by the work of Vaswani et al. [2017] on attention that has enabled startling progress in language models this year [Radford et al., 2019]. They also see the potential for research toward HLAI being aligned with some practical applications: scene understanding, autonomous agents and intelligent devices, autonomous driving and creative design.

### 2.4 Strengths and Limitations of Existing Work

Here, we briefly discuss strengths and limitations in order to motivate our call for an updated mapping of the HLAI technical landscape. Regarding the 2009 workshop, we think the systematic effort to be a significant strength, and notable as the only such effort. However, we believe that progress over the past decade has made this mapping less relevant. We see further limitations in the roadmapping method used, as it did not provide a roadmap but only a nominal mapping of milestones. We also are discouraged by the results due to the pessimism of organizers regarding them [Goertzel 2014, 2016].

We are less critical of the two more recent articles on paths toward HLAI [Mikolov et al., 2016; Lake et al., 2017]. We find each of them to constructively contribute to the conversation and to remain relatively valid despite continued rapid progress in the past several years. Lake et al. stands apart in its comprehensiveness, greater attention to technical elements and in its resilience to open commentary. Yet Mikolov et al. stands apart also, being the only proposed end-to-end method for training an HLAI agent. However, neither Lake et al. nor Mikolov et al. used well established scenario planning techniques to systematically enumerate plausible future scenarios, paths or technologies. Moreover, they both failed to create a directed graph or any visualization illustrating their proposed roadmap or ingredients' interactions. Consequently, and based on our familiarity with the scenario planning literature and our resulting understanding of its value to organizations and in multi-stakeholder situations, we feel strongly that an updated mapping should be conducted using these techniques to create a structured model of the existing knowledge and to uncover any blind spots or other potential missing pieces in our understanding of paths to HLAI.

## 3  Methods for Mapping

The methods mentioned for the mapping developed at the 2009 workshop used technology roadmapping. Here, we discuss these techniques briefly, but spend more time on two alternative techniques: scenario panning and online digital platforms. We believe that these alternative methods are promising for generating more useful results from a mapping.

### 3.1 Technology Roadmapping

Technology roadmapping is a widely used and flexible technology forecasting technique that is useful in supporting strategic and long-term planning [Phaal et al., 2004]. It often relies on a structured and graphical technique for exploring and communicating the future scenarios and is generally thought to consist of three phases: a preliminary phase, the development of the roadmap and a follow-up phase [Garcia & Bray 1997]. Technology roadmapping can be thought of in one of either two perspectives: an organizational perspective or a multi-organizational perspective.

Generally, there are eight different types of technology roadmaps including those for: a) product planning b) service/capability planning c) strategic planning d) long-range planning e) knowledge asset planning f) program planning g) process planning and h) integration planning [Phaal et al., 2004]. Furthermore, there are eight different types of graphical formats for roadmapping: a) multiple layers b) bars c) tables d) graphs e) pictorial representations f) flow charts g) single layer and h) text. Given the broad range of types and graphical formats, as well as the core three phases of roadmapping, it is easy to understand how the method is flexible. In application, these techniques are frequently molded to fit the specific case.

### 3.2 Scenario Planning

Scenario planning is a particularly powerful tool for understanding uncertainty and strategic planning in technology development. The techniques used for scenario building help teams to expand their thinking and commonly lead to uncovering blind spots and plausible futures that they would have otherwise not imagined. Scenario planning is also very flexible, and there is a significant amount of literature on the large range of techniques.

Scenario planning was first developed at the Rand Corporation in the 1960s [Bradfield et al., 2005]. It is traditionally associated with qualitative analysis, notably through what is widely known as the 'Shell method,' common for strategic planning in organizations. Due to the sustained success of Royal Dutch Shell in using the method (since the 1970s), it has been widely espoused in industry [Schoemaker 1995]. Scenario planning has been used widely for quantitative forecasting since the 1970s, and hybrid approaches, combining both qualitative and quantitative aspects, have seen widespread use since the 1980s [Amer et al., 2013]. Here, we consider only a small subset of scenario planning techniques relevant to mapping the HLAI technical landscape.

After a thorough literature review of scenario planning methods, we have identified a class of techniques that we refer to as *scenario mapping* techniques. These techniques all share the two significant properties that 1) there is no limit on the number of scenarios they can accommodate and 2) they represent the scenarios as networks with directed graphs. When represented as graphs, the nodes are scenarios that are connected by edges, which represent causal relations between the scenarios. These properties of the techniques are starkly different from the vast majority of scenario planning techniques[1] [Amer et al., 2013; Bradfield et al., 2005].

The first of these techniques that we consider is scenario network mapping (SNM) [List 2005]. SNM is a qualitative technique that was developed in order to accommodate more scenarios than traditional methods (30-50 scenarios is typical for SNM; 2-8 scenarios is typical for traditional methods). The technique is commonly more concerned with the network structure of the mapping than the actual scenarios generated. SNM uses multiday workshops of roughly 20 participants for the scenario building process, which focuses on generating clusters of scenarios rather than independent scenarios [List 2007]. This is a lesser known technique, but we believe it has significant potential due in large part to its incorporation of the holonic principle. The holonic principle means that each scenario exists simultaneously as a whole, as a part of a larger system and as a system comprised of smaller parts. An extension of SNM has demonstrated excellent results for forecasting complex networks of scenarios involving interactions between innovation at the micro and macro levels[2] [Gaziulusoy et al., 2013]. We believe that similar extensions[3] of this technique may be suitable for mapping HLAI.

---

[1] We note that the scenario mapping techniques identified aren't typically used for mapping technical components of emerging technologies. Rather, they're used for understanding the dynamic systems involved in the emergence of advanced technologies, e.g. economic factors, market forces, the rate of research progress in related fields, etc. However, HLAI research is in a class unto itself [Gruetzemacher 2018] in that it is attempting to create an artificial mind, and consequently, technology roadmapping fails. Despite their common use for planning problems, scenario mapping techniques are well suited for mapping the paths to HLAI. Technology roadmapping,,while popular, failed in its previous attempt for mapping HLAI [Goertzel 2014, 2016]. We explore this further in the discussion.

[2] Figure 7 in Gaziulusoy et al. [2013] gives a good example of how these mappings can be visualized. We don't expect a mapping of HLAI to take that shape, but such things are difficult to know beforehand as the method is intended to illuminate blind spots.

[3] Gaziulusoy developed a workshop technique specifically for sustainable organizations. It adapted List's methodology and used

The second of these techniques is cognitive maps, also a qualitative technique, which was first proposed by Axelrod in the 1970s as a way to represent social scientific knowledge and stakeholder beliefs in the form of a directed graph [Axelrod 2015; Amer et al., 2013]. However, Axelrod's use of the term was derivative, and its true origin was from cognitive psychology [Tolman 1948]. Lake et al. [2017] even refer to "cognitive maps" in advocating model-based reinforcement learning for modeling the way this notion of cognitive maps is thought to be used by the brain for planning. In the past decade they have come to be used for the purposes of scenario planning, but in this context are commonly referred to as causal maps [Goodier et al., 2010; Soetanto et al. 2011]. They are known to be effective for developing scenarios in complex multi-organizational cases. We think they are promising for understanding complex systems and see great potential for use in mapping cognitive processes in artificial agents.

Cognitive maps have not received much attention in scenario planning literature when compared to fuzzy cognitive maps (FCMs), which are simply a computable extension of cognitive maps that incorporate fuzzy logic [Amer et al., 2013]. First proposed in the 1980s, they are thought to be better for integrating expert, stakeholder and indigenous knowledge by enabling the development of scenarios that can aid in linking quantitative with qualitative storylines [Jetter & Schweinfort 2011]. While widely used for scenario planning, they have a wide range of other applications across numerous disciplines that generally involve complex modelling and decision-making tasks, e.g., online privacy management, decision support, knowledge representation and robotics [Papageorgiou 2013]. FCMs are weighted directed graphs with nodes that are fuzzy and representative of scenarios, or concepts, and with edges that represent causal relations. FCMs can be used to generate quantifiable forecasts, but their most significant feature may be their ability to integrate a wide variety of information types including subjective expert knowledge as well as technology, innovation and economic indicators. These properties can likely be extended to other scenario mapping techniques, or perhaps FCMs can be adapted for mapping the HLAI technical landscape.

### 3.3 Platforms and the Crowd

Crowdsourcing and platforms are very powerful tools for organizations in today's digital world [McAfee and Brynjolfsson, 2017]. We look to harness these tools in an attempt to address the challenges of HLAI mapping. The proposals here are not techniques, but rather means of eliciting expert opinion using powerful technologies. In combination with scenario analysis, we believe they can be used productively to improve future HLAI mapping efforts.

Perhaps the most straightforward way they could be used is simply for eliciting expert opinion via the Delphi technique. The Delphi technique could be beneficial to HLAI research as it typically incorporates the opinion of experts from a wide range of disciplines. A widely used proprietary scenario planning technique uses the Delphi method at a large scale to elicit expert opinion, i.e., up to 500 experts [Amer et al., 2013]. In order to use the Delphi technique at such scales, a specialized platform is needed. Commercial platforms are available for this purpose, and we believe that such platforms would be well suited for HLAI mapping efforts.

An alternate way to leverage the power of both platforms and the crowd is through the development of an online platform for user content creation. This can be easily done by using the MediaWiki open source code to create and launch a wiki for the purpose of technical discussions on HLAI development. All known architectures for HLAI could be entered and described in this HLAI wiki. Submodules and other components could also be described and linked to as appropriate. If the platform was populated thoroughly, or better, if it was widely used, then the data and structure of the network could be useful for developing a mapping. This utility could come either directly, in establishing a taxonomy of different types of HLAI, or indirectly, through further scenario analysis.

There likely exist other ways in which the power of platforms and the crowd can be exploited to improve HLAI mappings. Whether by the two methods mentioned here or one unmentioned, we think that it is appropriate for the HLAI research community to leverage the full potential of today's digital technology for furthering its own research. We think that using such platforms may also be a way to combine knowledge from the AI safety and HLAI research communities, who each spend substantial time thinking about HLAI.

## 4 Discussion

The HLAI technical landscape mapping from the 2009 workshop is the only systematic mapping of HLAI. However, results over the past decade in games and video games place the relevance these milestones into question [Vinyals et al., 2019; Ecoffet et al., 2019; Silver et al., 2017; Mnih et al., 2015]. Moreover, an advancement in attention mechanisms for deep nets [Vaswani et al. 2017] has led to language models that are markedly superior to the previous generation [Radford et al., 2019]. These language models are capable of elementary reading comprehension tasks and impressive text generation on common topics without tweaking or fine tuning for specific tasks. The advances in video games call into question 43.75% of the map depicted in Figure 1, while the advances in language modeling call into question substantially more, putting the total figure at arguably[4] over 50%.

---

workshopping techniques suitable for the context to improve outcomes for the specific use case. The same could be done in this case.

[4] We wish to note that it would be extreme to consider these numbers as realistic indicators of progress toward HLAI over the past decade. However, it would be equally as extreme to entirely discount them. We include them here simply to demonstrate that substantial progress has likely been made toward HLAI. We find it interesting to note that Gruetzemacher [2019] has recently shown

After a decade, and given the progress during that time, it seems appropriate to consider updating the existing mapping of the HLAI technical landscape. However, our call to update the HLAI mapping is not actually an updating at all – it is an effort to create a roadmap of the paths to HLAI that the 2009 workshop was unable to do. Having learned from their mistakes, we plan to revisit the challenge using the more powerful and rigorous techniques that we outlined in Sections 3.2, possibly informed by the techniques outlined in Section 3.3. We believe that in doing so we can create a graphical representation of the complex set of possible scenarios and technologies leading to HLAI. Thus, our roadmap may look more like the map from a sprawling network of suburban neighborhoods than an interstate, but it will be informative and offer new insights, some of which could be significant.

The best evidence for the use of these new techniques lies in the unsuccessful attempt of the 2009 workshop in developing a roadmap to HLAI [Goertzel 2014]. Upon failing to create a straightforward roadmap, the disappointed organizers proposed an analogy for mapping HLAI that we find appropriate: HLAI is like the peak of a mountain range and there are numerous paths up, the easiest of which is hard to determine until you reach the top [Goertzel 2016]. Routes up a mountain following ridges, cracks, aretes dihedrals, gullies and trails, all of which frequently intersect and can be mapped in the traditional sense. The only difference between this and a technology roadmap, is that it resembles a rural network of roads as opposed to a freeway. For HLAI, we believe there is value in a map of any sort, even if it doesn't proceed directly to the finish like a raceway. Thus, we feel very strongly that the conclusion from the only previous systematic HLAI mapping effort only underscores the need to test the methods proposed in this paper.

Although the plans laid out by Mikolov et al. [2016] and Lake et al. [2017] are strong guides for future research, they lack the benefits of using an objective and systematic technique such as those proposed. We in no way mean to diminish their brilliant contributions, rather, we wish to explicate the value of using a systematic method to build on it. If their frameworks were used as a starting point for the mapping process, then the result of the process could only enhance their work. Moreover, given the progress in the past two to three years, it is probable that simply updating these frameworks could yield progress due to the increases in collective knowledge – we likely have a better understanding of how to create HLAI than we did only a couple of years ago. It is also likely that blind spots will be illuminated through the use of scenario building exercises among experts, and it is possible that the results of a carefully crafted effort to map the paths to HLAI could significantly alter our perspectives on the existing frameworks.

The previous paragraphs have emphasized the possible positive impacts from our suggestions because we truly see little downside. At worst, we attain no real benefit to our understanding, but we would still gain an updated mapping that includes a directed graph modeling multiple frameworks simultaneously. We think that this risk is worth the effort because the possible gains to HLAI research could be substantial.

## 5 Conclusion

We have reviewed three frameworks for progress toward HLAI, including the sole previous systematic mapping of the HLAI technical landscape. We have considered the strengths and weaknesses of these previous studies, and we have proposed two alternative techniques[5] that are suitable for contributing to an updated HLAI mapping. In the discussion we considered the shortcomings of the previous methods, and explained in depth how the two new proposed techniques can sufficiently address them to provide the first true roadmap to HLAI. We intend for this paper to have the same effect as Goertzel et al. [2009] in motivating efforts to conduct a systematic mapping of the paths to HLAI, and we strongly encourage discussion and debate among the HLAI research community to initiate such efforts.

## Acknowledgments

We would like to thank Zoe Cremer, James Fox and Christy Manning for their comments, suggestions and assistance in various aspects of the development of this text.

## References

[Adams et al., 2012] Adams, S., I. Arel, J. Bach, R. Coop et al. "Mapping the landscape of human-level artificial general intelligence." AI magazine 33, no. 1 (2012): 25-42.

[Amer et al., 2013] Amer, M., T.U. Daim, and A. Jetter. "A review of scenario planning." Futures 46 (2013): 23-40.

[Axelrod 2015] Axelrod, Robert, ed. Structure of decision. Princeton university press, 2015.

[Baum et al., 2011] Baum, Seth D., Ben Goertzel, and Ted G. Goertzel. "How long until human-level AI? Results from an expert assessment." Technological Forecasting and Social Change 78, no. 1 (2011): 185-195.

[Botvinick et al. 2017] Botvinick, Matthew, David GT Barrett, Peter Battaglia, Nando de Freitas et al. "Building machines that learn and think for themselves: Commentary on lake et al., behavioral and brain sciences, 2017." arXiv preprint arXiv:1711.08378 (2017).

[Bradfield et al., 2005] Bradfield, Ron, George Wright, George Burt et al. "The origins and evolution of scenario techniques in long range business planning." Futures 37, no. 8 (2005): 795-812.

---

HLAI researchers to give significantly earlier and more precise predictions for transformative AI scenarios and for HLAI. For 2009 predictions from HLAI researchers, see in Baum et al. [2011].

[5] SNM appears to be the preferable technique for the task discussed here due to it's adherence to the holonic principle. Other scenario mapping techniques could be useful for other purposes in a broader forecasting framework for AI [Gruetzemacher 2019].

[Ecoffet et al., 2019] Ecoffet, Adrien, J. Huizinga, J. Lehman et al. "Go-Explore: a New Approach for Hard-Exploration Problems." arXiv preprint arXiv:1901.10995 (2019).

[Garcia & Bray 1997] Garcia, Marie, and Olin Bray. Fundamentals of technology roadmapping. No. SAND-97-0665. Sandia National Labs., Albuquerque, NM, 1997.

[Gziulusoy et al., 2013] Gaziulusoy, A. İdil, Carol Boyle, and Ron McDowall. "System innovation for sustainability: a systemic double-flow scenario method for companies." Journal of Cleaner Production 45 (2013): 104-116.

[Goertzel et al., 2009] Goertzel, Ben, Itamar Arel, and Matthias Scheutz. "Toward a roadmap for human-level artificial general intelligence" Artificial General Intelligence Roadmap Initiative 18 (2009): 27.

[Goertzel 2014] Goertzel, B. "Ten Years to the Singularity If We Really, Really Try." (2014).

[Goertzel 2016] Goertzel, Ben. The AGI Revolution. Humanity+ Press, 2016.

[Goodier et al., 2010] Goodier, Chris, Simon Austin, Robby Soetanto, and Andrew Dainty. "Causal mapping and scenario building with multiple organisations." Futures 42, no. 3 (2010): 219-229.

[Gruetzemacher 2018] Gruetzemacher, R. "Rethinking AI Strategy & Policy as Entangled Super Wicked Problems." 2018.

[Gruetzemacher 2019] Gruetzemacher, R. "A Holistic Framework for Forecasting Transformative AI." Forthcoming, 2019.

[Gruetzemacher et al., 2019] Gruetzemacher, R., D. Paradice, and K.B. Lee. "Forecasting Transformative AI: An Expert Survey." arXiv preprint arXiv:1901.08579 (2019).

[Jetter & Schweinfort 2011] A. Jetter and W. Schweinfort. "Building scenarios with Fuzzy Cognitive Maps." Futures 43, no. 1 (2011): 52-66.

[Laird 2009] Laird, John E., Robert E. Wray III, Robert P. Marinier III, and Pat Langley. "Claims and challenges in evaluating human-level intelligent systems." Proceedings of the 2nd Conf on AGI (2009). Atlantis Press, 2009.

[Laird & Wray, 2010] Laird, John E., and Robert E. Wray III. "Cognitive architecture requirements for achieving AGI." In 3d Conf on AGI (AGI-2010). Atlantis Press, 2010.

[Lake et al., 2017] Lake, Brenden, Ullman, Tomer, Tenenbaum, Joshua and Gershman, Samuel. "Building machines that learn and think like people." Behavioral and Brain Sciences 40 (2017).

[List 2005] List, Dennis. Scenario network mapping: the development of a methodology for social inquiry. Adelaide, SA, Australia: University of South Australia, 2005.

[List 2007] List, Dennis. "Scenario network mapping." Journal of Futures Studies 11, no. 4 (2007): 77-96.

[McAfee & Brynjolfsson, 2017] McAfee, Andrew, and Erik Brynjolfsson. Machine, platform, crowd: Harnessing our digital future. WW Norton & Company, 2017.

[Mikolov et al., 2016] Mikolov, Tomas, Armand Joulin, and Marco Baroni. "A roadmap towards machine intelligence." In International Conference on Intelligent Text Processing and Computational Linguistics, pp. 29-61. Springer, Cham, 2016.

[Mnih et al., 2015] Mnih, Volodymyr, Koray Kavukcuoglu, David Silver, Andrei A. Rusu et al. "Human-level control through deep reinforcement learning." Nature 518, no. 7540 (2015): 529.

[Moon 2007] Moon, Peter. "Three Minutes with Steve Wozniak." *PC World*. July (2007).

[Nilsson 2005] Nilsson, Nils. "Human-level artificial intelligence? Be serious!" AI magazine 26, no. 4 (2005): 68-68.

[Papageorgiou 2013] Papageorgiou, Elpiniki I., ed. Fuzzy cognitive maps for applied sciences and engineering: from fundamentals to extensions and learning algorithms. Vol. 54. Springer Science & Business Media, 2013.

[Phaal et al., 2004] Phaal, Robert, Clare JP Farrukh, and David R. Probert. "Technology roadmapping—a planning framework for evolution and revolution." Technological forecasting and social change 71, no. 1-2 (2004): 5-26.

[Piaget 1964] Piaget, Jean. "Part I: Cognitive development in children: Piaget development and learning." Journal of research in science teaching 2, no. 3 (1964): 176-186.

[Radford et al., 2019] Radford, Alex, Wu, Jeffrey, Child, Rewon, Luan, David et al. "Language Models are Unsupervised Multitask Learners." *OpenAI Blog*, Feb (2019).

[Ring 1997] Ring, M. "CHILD: A first step toward continual learning." Machine Learning 28, no. 1 (1997): 77-104.

[Schoemaker 1995] Schoemaker, Paul JH. "Scenario planning: a tool for strategic thinking." Sloan management review 36, no. 2 (1995): 25-41.

[Silver et al., 2017] Silver, D., J. Schrittwieser, K. Simonyan, I. Antonoglou et al. "Mastering the game of go without human knowledge." Nature 550, no. 7676 (2017): 354.

[Soetanto et al., 2011] Soetanto, Robby, Andrew RJ Dainty, Chris I. Goodier, and Simon A. Austin. "Unravelling the complexity of collective mental models: a method for developing and analysing scenarios in multi-organisational contexts." Futures 43, no. 8 (2011): 890-907.

[Tolman 1948] Tolman, Edward C. "Cognitive maps in rats and men." Psychological review 55, no. 4 (1948): 189.

[Vaswani et al., 2017] Vaswani, Ashish, Noam Shazeer, Niki Parmar, Jakob Uszkoreit et al. "Attention is all you need." In NIPS proceedings, pp. 5998-6008. 2017

[Vinyals et al., 2019] Vinyals, Oriol, Igor Babuschkin, Junyoung Chung, Michael Mathieu et al. AlphaStar: Mastering the Real-Time Strategy Game StarCraft II. DeepMind Blog. (2019).

[Vygotsky 1978] Vygotsky, Lev. "Interaction between learning and development." Readings on the development of children 23, no. 3 (1978): 34-41.